\documentclass[runningheads]{llncs}

\usepackage{eccv}

\usepackage{eccvabbrv}

\usepackage{graphicx}
\usepackage{booktabs}

\usepackage{makecell}
\usepackage{tabularx}
\usepackage[table]{xcolor}
\usepackage{multirow}

\usepackage[accsupp]{axessibility}  

\usepackage{booktabs}
\usepackage{siunitx}
\newcolumntype{Y}{>{\centering\arraybackslash}p{0.88cm}}

\usepackage[pagebackref,breaklinks,colorlinks,citecolor=eccvblue]{hyperref}
\usepackage{hyperref}

\usepackage{orcidlink}

\usepackage{pifont}
\newcommand{\cmark}{\ding{51}}
\newcommand{\xmark}{\ding{55}}

\begin{document}

\title{Semantic-Aware Reconstruction Error for Detecting AI-Generated Images}


\author{
Ju Yeon Kang\inst{1}\textsuperscript{*}
Jaehong Park\inst{1}\textsuperscript{*}
Semin Kim\inst{1}
Ji Won Yoon\inst{2}\textsuperscript{†}
Nam Soo Kim\inst{1}\textsuperscript{†}
}

\authorrunning{Kang et al.}

\institute{
Seoul National University, Seoul, South Korea \\
\email{\{jykang, jhpark, smkim21\}@hi.snu.ac.kr, nkim@snu.ac.kr} \and
Chung-Ang University, Seoul, South Korea \\
\email{jiwonyoon@cau.ac.kr}
}

\begingroup
\renewcommand{\thefootnote}{}
\footnotetext{* Equal contribution.}
\footnotetext{† Corresponding author.}
\endgroup

\maketitle

\begin{abstract}
    Recently, AI-generated image detection has gained increasing attention, as the rapid advancement of image generation technologies has raised serious concerns about their potential misuse. While existing detection methods have achieved promising results, their performance often degrades significantly when facing fake images from unseen, out-of-distribution~(OOD) generative models, since they primarily rely on model-specific artifacts and thus overfit to the models used for training. To address this limitation, we propose a novel representation, namely Semantic-Aware Reconstruction Error~(SARE), that measures the semantic difference between an image and its caption-guided reconstruction. The key hypothesis behind SARE is that real images, whose captions often fail to fully capture their complex visual content, may undergo noticeable semantic shifts during the caption-guided reconstruction process. In contrast, fake images, which closely align with their captions, show minimal semantic changes. By quantifying these semantic shifts, SARE provides a robust and discriminative feature for detecting fake images across diverse generative models. Additionally, we introduce a fusion module that integrates SARE into the backbone detector via a cross-attention mechanism. Image features attend to semantic representations extracted from SARE, enabling the model to adaptively leverage semantic information. Experimental results demonstrate that the proposed method achieves strong generalization, outperforming existing baselines on benchmarks including GenImage and ForenSynths. We further validate the effectiveness of caption guidance through a detailed analysis of semantic shifts, confirming its ability to enhance detection robustness. The code and pretrained models will be released upon acceptance.
    \keywords{AI-Generated Image Detection \and Diffusion Models}
\end{abstract}



\begin{figure}[t]
    \centering
    \includegraphics[width=\linewidth]{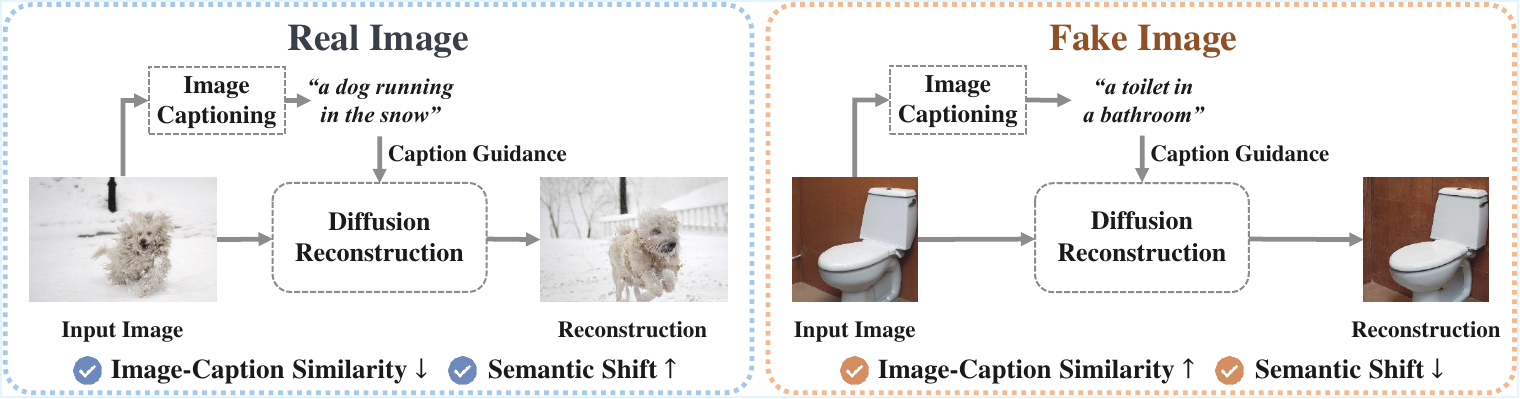}
    \caption{Comparison of caption-guided reconstructions for real and fake images. Real images, whose captions often fail to fully capture their complex visual content, undergo noticeable semantic shifts during caption-guided reconstruction. In contrast, fake images, which align closely with their captions, tend to exhibit minimal semantic changes.}
    \label{fig:intro_fig}
\end{figure}

\section{Introduction}
\label{sec:intro}

In recent years, image generation technologies, such as Generative Adversarial Networks (GANs)~\cite{gan,cyclegan,biggan,progan} and Diffusion Models (DMs)~\cite{ddpm,ddim,sd,glide}, have made remarkable progress, enabling the synthesis of highly realistic images that are often indistinguishable from real images. This realism has raised growing concerns about potential misuse, particularly in generating harmful or deceptive content~\cite{ferreira2020review,juefei2022countering}. To address these risks, developing reliable methods for detecting AI-generated images has become increasingly important.

A common approach in existing detection methods is to train a binary classifier using real and fake images sourced from a finite set of generative models available during training~\cite{bayar2016deep,cnnspot,gramnet,dire}.
While these detectors typically exhibit strong performance when test images are generated by the same models used for training, their performance often drops significantly in real-world scenarios, where they inevitably encounter fake images from unseen generative models that are not included in the training dataset~\cite{zhang2019detecting,luo2021generalizing,yan2023ucf}.
To ensure robustness in practical deployment, it is essential to develop detection methods that can generalize effectively to such unseen and out-of-distribution~(OOD) generative models.

Recent studies have proposed several strategies to address the generalization challenges inherent in generated image detection.
These strategies include training methods such as reconstruction-based learning~\cite{dire,lare,fire} and data augmentation~\cite{drct}, as well as architectural approaches~\cite{univfd,lasted,c2pclip} that leverage a large pre-trained model like CLIP~\cite{vanillaclip}.
Despite these advances, the robustness of existing methods remains limited, as they primarily focus on identifying visual artifacts introduced during the generative process~\cite{FreDect,cnnspot,dire,drct}.
Due to the distinct characteristics of different generative models, such artifacts are inherently model-specific and fail to generalize across diverse models~\cite{luo2021generalizing,corvi2023detection,univfd}.
As a result, approaches that rely on these artifacts tend to overfit to the models used for training, which leads to degraded performance in OOD scenarios.

To overcome these limitations, we explore a fundamental property commonly observed in fake images.
Prior work~\cite{defake} has shown that the similarity between fake images and captions generated by an image-captioning model is typically higher than that of real images.
Real images contain complex, fine-grained details that short captions cannot cover, whereas fake images include only the elements explicitly specified in the user's text prompt.
Inspired by this observation, we hypothesize that the relationship between an image and its caption reflects a general characteristic of fake images, providing a robust signal for detection across diverse generative models.

In this paper, we propose Semantic-Aware Reconstruction Error (SARE), a novel representation for detecting AI-generated images that measures the semantic difference between an image and its reconstruction. 
Specifically, we introduce a caption-guided reconstruction pipeline to effectively leverage the relationship between an image and its caption in the detection process.
The key idea is that real images, which often exhibit low similarity to their captions, may undergo noticeable semantic shifts during caption-guided reconstruction.
In contrast, fake images, whose content is well captured by their captions, show minimal semantic shifts.
As shown in \cref{fig:intro_fig}, the real image is reconstructed into a noticeably different dog since the caption provides only a coarse description (\eg, “a dog running in the snow”) without capturing fine details such as the dog's breed, pose, or background.
Conversely, the fake image of a toilet remains largely unchanged after reconstruction, as its content can be sufficiently described by a simple caption.
By capturing these fundamental differences between real and fake images, SARE provides a discriminative and generalizable feature for detecting fake images across diverse generative models.
Additionally, we design a fusion module that integrates SARE into the backbone detector via a cross-attention mechanism.
The original image features attend to the semantic representations extracted from SARE, allowing the model to adaptively incorporate semantic information.

We validate the effectiveness of SARE through extensive experiments on GenImage~\cite{genimage} and ForenSynths~\cite{cnnspot} datasets.
The proposed framework significantly improves the performance of the backbone model across both seen and unseen generators, achieving the best average results compared to existing detection methods.
The results demonstrate the robustness of SARE in OOD scenarios, confirming its strong generalization to diverse generative models. 

\section{Related Work}
\label{sec:related}

\subsection{Detection Based on Image Caption}
Caption-based detection methods explore the use of image captions as a cue for detecting generated images.
DE-FAKE~\cite{defake} finds that generated images tend to align more closely with their captions compared to real images.
Based on this observation, it adopts separate encoders for image and caption to exploit the relationship between them.
Following this direction, C2P-CLIP~\cite{c2pclip} proposes a method that injects category-level prompts to enhance detection performance.
LASTED~\cite{lasted} introduces a language-guided contrastive learning framework that leverages textual labels to improve generalization.

\subsection{Detection Based on Image Reconstruction}
Reconstruction-based detection methods utilize a pre-trained diffusion model to reconstruct the input image and analyze the differences between the original and reconstructed images.
DIRE~\cite{dire} introduces reconstruction error as the discriminative feature for detection, based on the assumption that fake images can be reconstructed more accurately than real images.
AEROBLADE~\cite{aeroblade} proposes a training-free approach for detecting images from latent diffusion models using autoencoder reconstruction errors.
DRCT~\cite{drct}, rather than relying on reconstruction error, treats reconstructed images as hard samples and adopts a contrastive learning framework to facilitate discriminative feature learning.

Despite these advances, the incorporation of caption conditioning into the reconstruction process remains largely unexplored. For example, FakeInversion~\cite{fakeinversion} employs caption guidance during DDIM inversion~\cite{ddim}, but performs the reconstruction without classifier-free guidance (CFG)~\cite{cfg}. Consequently, the caption does not induce semantic changes but instead serves only to stabilize the inversion process, as noted in prior works~\cite{nulltextinv,etainversion,videvo}. In contrast, our method applies CFG to explicitly induce caption-driven semantic shifts and leverages these shifts as the core signal for detection.

\begin{figure}[t]
    \centering
    \includegraphics[width=\linewidth]{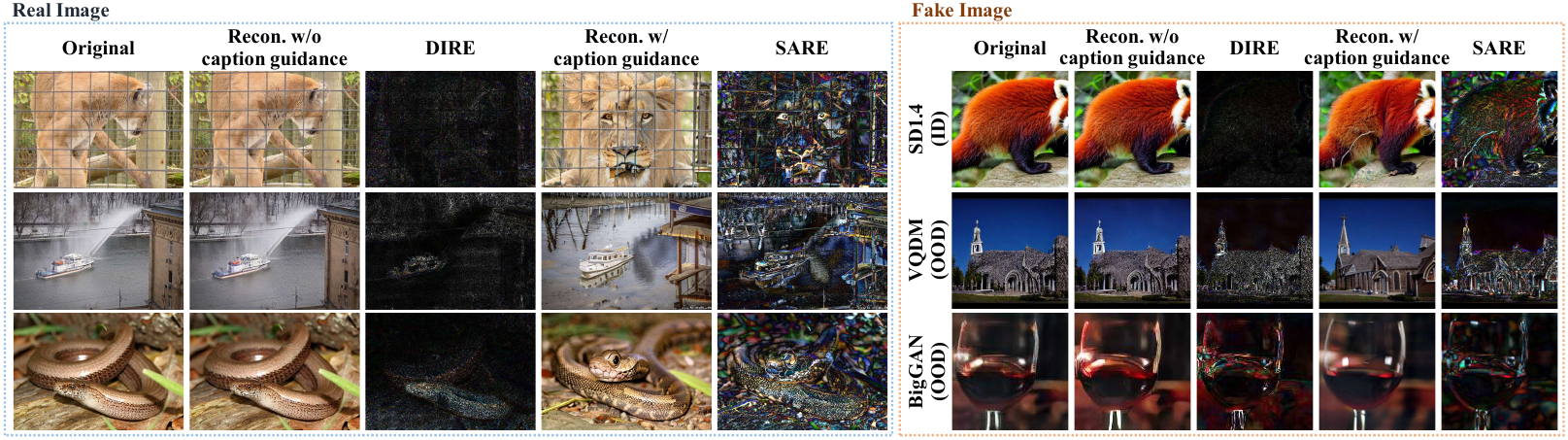}
    \caption{Examples from the GenImage dataset~\cite{genimage} and their corresponding DIREs~\cite{dire}, and SAREs. Images are reconstructed using Stable Diffusion v1~\cite{sd}. DIRE may produce larger errors for OOD fake images than for real images, contradicting its underlying assumption. In contrast, SARE consistently yields higher values for real images than for fake images. For clearer visualization, the pixel values of the error maps are scaled by 2.}
    \label{fig:motivation}
\end{figure}

\section{Proposed Method}
\label{sec:method}
                                            
\begin{figure*}[t]
  \centering
  \includegraphics[width=\linewidth]{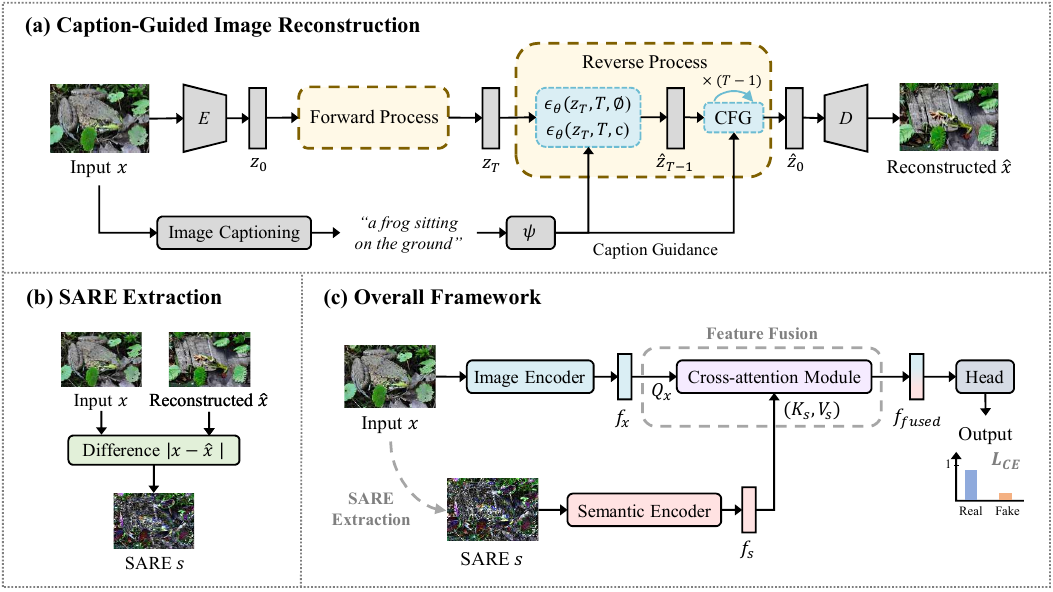}
  \caption{Overview of the SARE framework. Our method reconstructs the input image conditioned on its caption using the Stable Diffusion model~\cite{sd} with classifier-free guidance. SARE is computed as the difference between the input and reconstructed image, and is incorporated into the detection process through a cross-attention module that leverages image features as queries and SARE features as keys and values. The pixel values of the SARE are scaled by 2 for clearer visualization.}
  \label{fig:overall}
\end{figure*}




\subsection{Motivation}

Existing methods~\cite{FreDect,cnnspot,dire,drct, fakeinversion} for detecting fake images primarily rely on visual artifacts or traces left by the generative models.
A representative example is DIRE~\cite{dire}, which reconstructs the input image with a pre-trained diffusion model and leverages the pixel-wise reconstruction error as a discriminative feature for classification.
It is based on the assumption that fake images exhibit smaller reconstruction errors than real images, as both the original and reconstructed images belong to the same generative distribution and thus share similar visual patterns.
However, our empirical observation suggests that this assumption often does not hold in OOD scenarios, where fake images are synthesized by unseen generators that were not available during training.
As shown in \cref{fig:motivation}, when Stable Diffusion v1~\cite{sd} is used for reconstruction, fake images from unseen models such as ADM~\cite{adm} or BigGAN~\cite{biggan} produce much larger reconstruction errors~(\ie, DIREs), even exceeding those of real images.
These findings indicate that diverse generative models, including the reconstruction model and unseen generators, exhibit distinctive characteristics and traces.
Therefore, detection methods that rely on visual artifacts from a specific generation process may struggle to generalize in OOD scenarios.
This limitation highlights the need for more robust and generalizable detection cues that can perform reliably across diverse generators.

\subsection{Semantic-Aware Reconstruction Error}
\label{sec3.2}
We propose Semantic-Aware Reconstruction Error (SARE), a novel detection feature designed to enhance generalization in AI-generated image detection. The hypothesis of SARE is that the relationship between an image and its caption may reflect fundamental differences between real and fake images, and thus serve as a generalizable detection cue. SARE aims to effectively leverage this property by introducing a caption-guided reconstruction framework. Our proposed method consists of three main steps: (a) image captioning, (b) caption-guided image reconstruction, and (c) SARE extraction. An overview of the SARE framework is illustrated in \cref{fig:overall}.

\subsubsection{Image Captioning.}
For a given image $x$, we utilize a pre-trained image captioning model to generate a descriptive caption $C$.
This caption $C$ is used as the text condition for the subsequent reconstruction process.

\subsubsection{Caption-Guided Image Reconstruction.}
Given the caption $C$, we reconstruct the input image $x$ by using a pre-trained text-conditional diffusion model.
Specifically, we leverage the Stable Diffusion model~\cite{sd} with classifier-free guidance~\cite{cfg}.
The input image $x$ is first encoded into a latent representation $z_0$ using the Variational Autoencoder (VAE) encoder~\cite{vae}.
The forward process then adds Gaussian noise to ${z}_0$ following a predefined noise schedule.
The noisy latent at a given timestep $t$ is computed as:
\begin{equation}
z_t = \sqrt{\bar{\alpha}_t} z_0 + \sqrt{1 - \bar{\alpha}_t} \epsilon,
\end{equation}
where $\epsilon \sim \mathcal{N}(0,\mathbf{I})$, and $\bar{\alpha}_t = \prod_{s=1}^{t} {\alpha}_s$.
The \textit{strength} parameter determines the amount of noise added during reconstruction.
The number of forward diffusion steps is set to $T = \lfloor \textit{strength} \times T_{\max} \rfloor$, where $T_{\max}$ is the total number of diffusion steps.
\label{paragraph:Caption-guided Image Reconstruction}

Starting from the noisy latent $z_T$, the reverse process aims to obtain $\hat{z}_0$ through an iterative denoising process conditioned on the caption $C$.
At each denoising step, the noise prediction network $\epsilon_{\theta}(z_t,t,c)$ estimates the noise $\epsilon$, where $c=\psi(C)$ denotes the caption embedding obtained from the CLIP text encoder~\cite{vanillaclip}.
We adopt classifier-free guidance, which combines the conditional and unconditional noise predictions as follows:
\begin{equation}
\epsilon_{\theta}(z_t,t,c,\varnothing)=w\epsilon_\theta(z_t,t,c)+(1-w)\epsilon_\theta(z_t,t,\varnothing),
\end{equation}
where $w$ is the guidance scale and $\varnothing=\psi(\text{“”})$ denotes the null text embedding.
The denoising process using DDIM sampling~\cite{ddim} can be represented by:
\begin{equation}
z_{t-1} = \sqrt{\alpha_{t-1}} \frac{z_t - \sqrt{1 - \alpha_t} \, \epsilon_{\theta}(z_t,t,c,\varnothing)}{\sqrt{\alpha_t}} + \sqrt{1 - \alpha_{t-1}} \epsilon_t,
\label{eq:DDIM}
\end{equation}
where $\alpha_{t-1} = \frac{\bar\alpha_{t-1}}{\bar\alpha_{t}}$ and $\epsilon \sim \mathcal{N}(0,\mathbf{I})$, for $t=T,...,1$.
After $T$ denoising steps, the final latent $\hat{z}_0$ is obtained and decoded by the VAE decoder to produce the reconstructed image $\hat{x}$.

\subsubsection{SARE Extraction.}
Once we obtain the original image $x$ and the reconstructed image $\hat{x}$, we compute the SARE by measuring the difference between the two images.
SARE is defined as follows:
\begin{equation}
\text{SARE}(x, \hat{x}) = |x - \hat{x}|,
\end{equation}
where $| \cdot |$ denotes the absolute value. 
SARE quantifies the semantic changes introduced during the caption-guided reconstruction process. 
Since real images often contain complex visual details that cannot be fully reflected in their captions, their reconstructions result in noticeable semantic shifts. 
In contrast, fake images typically align closely with their captions and therefore tend to undergo minimal semantic changes. 
By capturing these differences between real and fake images, SARE can serve as a discriminative feature for robust detection across diverse generative models.

\subsection{Fusion Module}
We propose a fusion module to effectively integrate SARE into the detection process.
Given an input image $x$ and its corresponding SARE $s$, we extract the image feature $f_x$ and the semantic feature $f_s$ using the image encoder $E_x$ and the semantic encoder $E_s$, respectively:
\begin{equation}
    f_x = E_x(x),\ f_s = E_s(s).
\end{equation}
To obtain the fused feature $f_{fused}$, we employ a cross-attention mechanism by leveraging $f_x$ for query and $f_s$ for key and value as follows:
\begin{equation}
\begin{array}{c}
    Q_x = f_x W_Q, \; K_s = f_sW_K, \; V_s = f_sW_V, \\[6pt]
    f_{fused} = CrossAttn(Q_x, K_s, V_s),
\end{array}
\end{equation}
where $W_Q, W_K$, and $W_V$ are the linear projections for the query, key, and value, respectively.
This fused representation allows the model to incorporate semantic information as an additional cue.
Subsequently, $f_{fused}$ is passed through a fully connected layer that serves as the classification head, and the model is trained using binary cross-entropy loss.
\section{Experiment}
\label{sec:exp}


We first introduce the experimental settings and then present the results to evaluate our proposed method. Additional results and analyses are provided in the supplementary material.

\subsection{Experimental Settings}
\subsubsection{Datasets and Evaluation Metrics.}
We evaluated the performance of detection models using the GenImage~\cite{genimage} dataset, which is divided into 8 subsets.
Each subset consists of real images from ImageNet~\cite{imagenet} and fake images synthesized by a single generative model.
The generative models are Midjourney (MJ)~\cite{Mid}, Stable Diffusion v1.4\&v1.5 (SDv1.4\&v1.5)~\cite{sd}, ADM~\cite{adm}, GLIDE~\cite{glide}, Wukong~\cite{Wukong}, VQDM~\cite{vqdm}, and BigGAN~\cite{biggan}.
We used the training split from the SDv1.4 subset for training and randomly selected 10\% of it as a validation set, since GenImage does not provide an official validation split. The test splits from all subsets were used for evaluation.
For cross-dataset evaluation, we trained the models on the SDv1.4 subset of GenImage and evaluated them on the ForenSynths~\cite{cnnspot} test set.
The ForenSynths test set contains 11 subsets, where each subset comprises real images from the training data of a specific generative model and fake images produced by that model. The generative models in ForenSynths include ProGAN~\cite{progan}, StyleGAN~\cite{stylegan}, StyleGAN2~\cite{stylegan2}, BigGAN~\cite{biggan}, CycleGAN~\cite{cyclegan}, StarGAN~\cite{stargan}, GauGAN~\cite{gaugan}, CRN~\cite{crn}, IMLE~\cite{imle}, SITD~\cite{sitd}, SAN~\cite{san}, and Deepfake~\cite{deepfake}.
For evaluation metrics, we employed accuracy~(ACC) and the Area Under the ROC curve~(AUC). Accuracy was computed with a fixed threshold of 0.5, following the baseline settings~\cite{dire,drct}.

\begin{table}[t]

\caption{Accuracy (ACC, \%) comparisons of different detectors on the GenImage dataset~\cite{genimage}. Models are trained on the SDv1.4 subset and evaluated across 8 subsets. The best and second-best results are indicated in \textbf{bold} and \underline{underlined}, respectively.}
\label{main_acc}

\centering
\small
\setlength{\tabcolsep}{3pt}

\resizebox{\textwidth}{!}{
\begin{tabular}{lccccccccc}
\toprule
Method & MJ & SDv1.4 & SDv1.5 & ADM & GLIDE & Wukong & VQDM & BigGAN & Avg.(\%) \\
\midrule
GramNet~\cite{gramnet}  & 73.32 & 96.73 & 96.55 & 51.73 & 58.85 & 91.19 & 57.05 & 48.63 & 71.76 \\
Conv-B~\cite{convb}   & 84.59 & \textbf{100.00} & \textbf{99.91} & 52.86 & 57.14 & \textbf{99.88} & 58.77 & 50.01 & 75.40 \\
UnivFD~\cite{univfd}   & 89.56 & 96.94 & 96.56 & 57.20 & 71.12 & 95.03 & 68.67 & 57.83 & 79.11 \\
DIRE~\cite{dire}     & 51.03 & \underline{99.96} & \textbf{99.91} & 51.78 & 59.26 & \underline{99.79} & 50.18 & 50.88 & 70.35 \\
DE-FAKE~\cite{defake}  & 85.55 & 97.93 & 97.82 & 53.53 & 65.28 & 91.57 & 55.98 & 49.16 & 74.60 \\
NPR~\cite{npr}  & 89.69 & 90.54 & 90.56 & \underline{84.42} & 90.15 & 90.43 & 86.72 & \underline{90.65} & \underline{89.15} \\
AIDE~\cite{aide}     & 79.38 & 99.74 & 99.76 & 78.54 & \underline{91.82} & 98.65 & 80.25 & 66.88 & 86.88 \\
\midrule
DRCT~\cite{drct}     & \textbf{90.89} & 94.75 & 94.28 & 78.54 & 87.52 & 94.58 & \underline{90.12} & 79.76 & 88.81 \\
\rowcolor{gray!20}
SARE (ours)
         & \underline{90.32} & 97.21 & 97.04 & \textbf{84.47} & \textbf{93.55} & 97.05 & \textbf{93.66} & \textbf{92.05} & \textbf{93.17} \\
\bottomrule
\end{tabular}
}

\end{table}

\begin{table}[t]

\caption{AUC (\%) comparisons of different detectors on the GenImage dataset~\cite{genimage}. Models are trained on the SDv1.4 subset and evaluated across 8 subsets. The best and second-best results are indicated in \textbf{bold} and \underline{underlined}, respectively.}
\label{main_auc}

\centering
\small
\setlength{\tabcolsep}{3pt}

\resizebox{\textwidth}{!}{
\begin{tabular}{lccccccccc}
\toprule
Method & MJ & SDv1.4 & SDv1.5 & ADM & GLIDE & Wukong & VQDM & BigGAN & Avg.(\%) \\
\midrule
GramNet~\cite{gramnet} & 91.54 & 99.56 & 99.49 & 69.87 & 83.52 & 98.10 & 78.40 & 39.36 & 82.48 \\
Conv-B~\cite{convb}  & \textbf{99.54} & \textbf{100.00} & \underline{99.94} & 90.10 & 96.72 & \textbf{100.00} & 93.82 & 86.61 & 95.84 \\
UnivFD~\cite{univfd}  & 97.54 & 99.57 & 99.51 & 73.09 & 89.46 & 98.99 & 87.53 & 79.19 & 90.61 \\
DIRE~\cite{dire}    & 78.65 & \textbf{100.00} & \underline{99.94} & 71.45 & 90.42 & \underline{99.99} & 62.49 & 61.12 & 83.01 \\
DE-FAKE~\cite{defake} & 97.13 & 99.81 & 99.80 & 70.95 & 89.26 & 98.52 & 78.48 & 57.60 & 86.44 \\
NPR~\cite{npr} & \underline{97.67} & 99.10 & 99.14 & \underline{93.04} & \underline{98.50} & 98.39 & 93.62 & \textbf{98.67} & \underline{97.27} \\
AIDE~\cite{aide}    & 95.94 & 99.99 & \textbf{99.98} & 92.12 & \textbf{98.66} & 99.97 & \underline{96.54} & 92.04 & 96.91 \\
\midrule
DRCT~\cite{drct} & 96.91 & 99.64 & 99.52 & 88.47 & 94.61 & 99.42 & 96.44 & 90.30 & 95.66 \\
\rowcolor{gray!20}
SARE (ours)
     & 96.83 & 99.94 & 99.93 & \textbf{94.87} & 98.00 & 99.83 & \textbf{98.31} & \underline{97.51} & \textbf{98.15} \\
\bottomrule
\end{tabular}
}

\end{table}

\subsubsection{Implementation Details.}
To obtain reconstructed images for SARE and for the baseline models DIRE~\cite{dire} and DRCT~\cite{drct}, we used SDv1 as the reconstruction model.
For SARE and DE-FAKE~\cite{defake}, captions were generated using a pre-trained BLIP model~\cite{blip}.
During the reconstruction, we set the strength parameter to 0.5, the guidance scale to 7.5, and the maximum number of diffusion steps to 50, which results in 25 forward diffusion steps as described in Sec.~\ref{sec3.2}.
We adopted DRCT as the backbone detector, which utilizes CLIP:ViT-L/14~\cite{vanillaclip} as the image encoder.
For the semantic encoder, we employed a ResNet50 model~\cite{resnet}.
During training, we applied several augmentations, including random cropping, horizontal flipping, Gaussian noise injection, Gaussian blurring, random rotation, JPEG compression with random quality, random scaling, grid dropout, and brightness and contrast adjustments.
At test time, images were center-cropped to 224 $\times$ 224.
For SARE extraction, images were resized to 512 on the longer side before reconstruction, and the resulting SARE representations were fed into the encoder at a size of 224 $\times$ 224.
We trained the models for 17 epochs using the AdamW optimizer~\cite{adamw} with an initial learning rate of $1 \times 10^{-4}$ and selected the final model based on the best validation accuracy.
All baseline detectors, except for NPR~\cite{npr} and AIDE~\cite{aide}, were re-implemented under the same training settings described above.
For NPR and AIDE, we used the official pre-trained checkpoints.

\begin{table}[t]

\caption{Accuracy (ACC, \%) comparisons of different detectors under cross-dataset evaluation. All detectors are trained on the SDv1.4 subset of the GenImage dataset~\cite{genimage} and evaluated on the ForenSynths test set~\cite{cnnspot}. The best and second-best results are indicated in \textbf{bold} and \underline{underlined}, respectively.}
\label{ood_acc}

\centering
\small
\setlength{\tabcolsep}{2.7pt}
\resizebox{\textwidth}{!}{
\begin{tabular}{lccccccccccccc}
\toprule
Method &
\makecell{Pro-\\GAN} &
\makecell{Style-\\GAN} &
\makecell{Style-\\GAN2} &
\makecell{Big-\\GAN} &
\makecell{Cycle-\\GAN} &
\makecell{Star-\\GAN} &
\makecell{Gau-\\GAN} &
CRN & IMLE & SITD & SAN &
\makecell{Deep-\\Fake} &
Avg.(\%) \\
\midrule
GramNet~\cite{gramnet}  & 49.20 & 48.62 & 48.52 & 49.73 & 49.17 & 49.07 & 48.70 & 47.59 & 47.50 & 65.56 & 57.99 & 58.02 & 51.64 \\
Conv-B~\cite{convb}     & 54.66 & 50.92 & 50.01 & 52.50 & 50.04 & 49.47 & 50.19 & \textbf{49.94} & 52.50 & 62.50 & 66.44 & \textbf{80.19} & 55.78 \\
UnivFD~\cite{univfd}    & 67.96 & 55.81 & 52.04 & 68.55 & 69.23 & \underline{79.89} & 56.23 & 38.08 & 54.64 & 64.17 & 65.53 & \underline{60.56} & 61.06 \\
DIRE~\cite{dire}        & 50.06 & 50.04 & 50.00 & 49.88 & 49.96 & 50.05 & 49.97 & 49.42 & 49.58 & 54.17 & 73.29 & 52.56 & 52.42 \\
DE-FAKE~\cite{defake}   & 51.20 & 49.32 & 47.47 & 52.88 & 51.89 & 63.81 & 49.02 & 49.46 & 47.31 & 53.89 & 65.30 & 51.77 & 52.78 \\
NPR~\cite{npr}          & 58.11	& 65.51	& 61.65	& 58.48	& 73.77	& 63.76	& 53.55	& 48.79	& 48.79	& 51.67	& 58.68	& 49.92 & 57.72 \\
AIDE~\cite{aide}        & 69.27 & \underline{71.04} & \underline{72.53} & 77.20 & 74.37 & \textbf{80.24} & 64.35 & \underline{49.61} & \textbf{67.65} & \textbf{69.72} & 57.53 & 54.24 & 67.31 \\
\midrule
DRCT~\cite{drct}        & \underline{74.59} & 68.83 & 66.04 & \underline{83.10} & \textbf{93.11} & 62.23 & \underline{78.87} & 41.69 & 51.83 & \underline{66.67} & \underline{79.45} & 55.76 & \underline{68.51} \\
\rowcolor{gray!20}
SARE (ours)
         & \textbf{84.51} & \textbf{76.74} & \textbf{76.04} & \textbf{83.15} & \underline{90.95} & 59.65 & \textbf{81.28} & 46.65 & \underline{60.98} & 61.39 & \textbf{84.70} & 51.56 & \textbf{71.47} \\
\bottomrule
\end{tabular}
}

\end{table}

\begin{table}[t]

\caption{AUC~(\%) comparisons of different detectors under cross-dataset evaluation. All detectors are trained on the SDv1.4 subset of the GenImage dataset~\cite{genimage} and evaluated on the ForenSynths test set~\cite{cnnspot}. The best and second-best results are indicated in \textbf{bold} and \underline{underlined}, respectively.}
\label{ood_auc}

\centering
\small
\setlength{\tabcolsep}{2.7pt}

\resizebox{\textwidth}{!}{
\begin{tabular}{lccccccccccccc}
\toprule
Method &
\makecell{Pro-\\GAN} &
\makecell{Style-\\GAN} &
\makecell{Style-\\GAN2} &
\makecell{Big-\\GAN} &
\makecell{Cycle-\\GAN} &
\makecell{Star-\\GAN} &
\makecell{Gau-\\GAN} &
CRN & IMLE & SITD & SAN &
\makecell{Deep-\\Fake} &
Avg.(\%) \\
\midrule
GramNet~\cite{gramnet}  & 49.00 & 44.69 & 46.90 & 50.76 & 55.90 & 48.46 & 34.39 & \underline{49.90} & 39.23 & 75.13 & 70.14 & 63.88 & 52.37 \\
Conv-B~\cite{convb}     & 75.08 & 76.94 & 72.15 & 77.46 & 52.96 & 38.18 & 62.23 & 44.21 & \underline{85.55} & \underline{86.54} & \textbf{98.62} & \textbf{87.58} & 71.46 \\
UnivFD~\cite{univfd}    & 80.55 & 67.47 & 62.97 & 84.46 & 93.56 & 89.32 & 80.03 & 29.51 & 57.22 & 74.73 & 75.08 & 67.96 & 71.91 \\
DIRE~\cite{dire}        & 55.59 & 52.28 & 52.44 & 45.27 & 47.96 & 51.99 & 45.38 & 43.67 & 62.66 & \textbf{93.88} & \underline{98.45} & \underline{84.30} & 61.16 \\
DE-FAKE~\cite{defake}   & 55.57 & 49.71 & 44.33 & 70.09 & 74.58 & 71.15 & 43.10 & \textbf{51.76} & 46.21 & 51.93 & 77.38 & 51.11 & 57.24 \\
NPR~\cite{npr}          & 73.12	& \underline{87.40}	& 75.13	& 64.77	& 78.4	& \textbf{99.49}	& 51.80	& 41.58	& 49.80	& 55.15	& 72.32	& 55.54 & 67.04 \\
AIDE~\cite{aide}        & 84.24	& \textbf{87.71} & \textbf{90.58} & 85.67 & 94.55 & 91.06 & 72.56 & 47.57 & \textbf{93.82} & 70.03 & 77.09 & 59.72 & \underline{79.55} \\
\midrule
DRCT~\cite{drct}        & \underline{89.08} & 77.05 & 73.24 & \textbf{92.74} & \textbf{98.31} & \underline{95.93} & \underline{88.23} & 29.35 & 68.56 & 79.39 & 88.75 & 70.01 & 79.22 \\
\rowcolor{gray!20}
SARE (ours)
         & \textbf{93.12} & 85.97 & \underline{83.60} & \underline{92.11} & \underline{96.09} & 94.84 & \textbf{90.45} & 47.71 & 79.08 & 77.64 & 92.55 & 77.72 & \textbf{84.24} \\
\bottomrule
\end{tabular}
}

\end{table}

\subsection{Comparisons to Existing Detectors}
\label{main}
\cref{main_acc} and \cref{main_auc} report the accuracies and AUC scores of different detection methods on the GenImage dataset.
We compared our method with several detectors, including GramNet~\cite{gramnet}, Conv-B~\cite{convb}, UnivFD~\cite{univfd}, DIRE, DE-FAKE, NPR, AIDE, and DRCT.
All models were trained on the SDv1.4 subset.
The results show that compared to DRCT, our method improves the average accuracy by 4.36$\%$, and the average AUC by 2.49$\%$, which indicates that integrating SARE effectively enhances the detection performance.
Notably, our method achieves the highest average accuracy of 93.17$\%$ and AUC of 98.15$\%$, outperforming all other detection approaches.
While all the detectors show strong performance on SDv1.4, SDv1.5, and Wukong subsets, their performance tends to degrade significantly on other subsets like ADM, GLIDE, VQDM, and the non-diffusion model BigGAN.
Our method maintains consistently high performance across all subsets, demonstrating robust generalization to diverse OOD generative models.
Moreover, the proposed method outperforms DE-FAKE, suggesting that SARE leverages the relationship between an image and its caption more effectively than directly comparing image and caption embeddings obtained from CLIP.

\begin{figure*}[t]
    \centering
    \begin{subfigure}[b]{\linewidth}
        \centering
        \includegraphics[width=\linewidth]{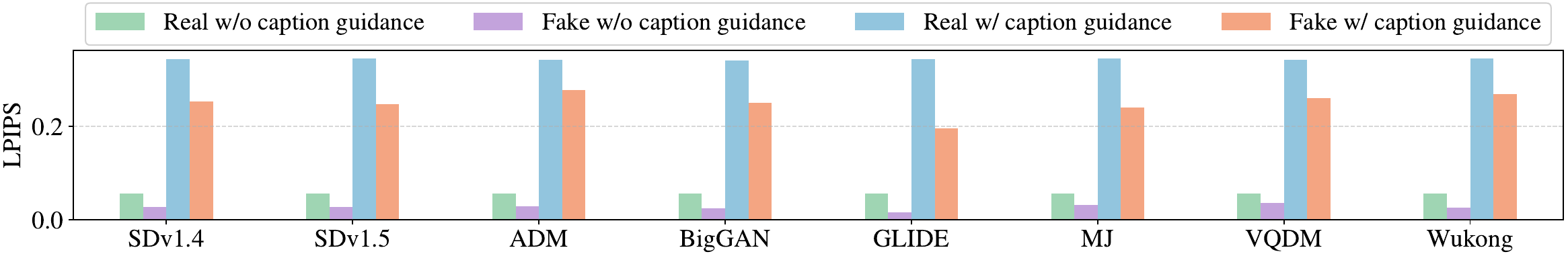}
        \caption{Average LPIPS scores between original images and their reconstructions.}
        \label{lpips_plot}
    \end{subfigure}
    \vspace{0.1em}
    \begin{subfigure}[b]{\linewidth}
        \centering
        \includegraphics[width=\linewidth]{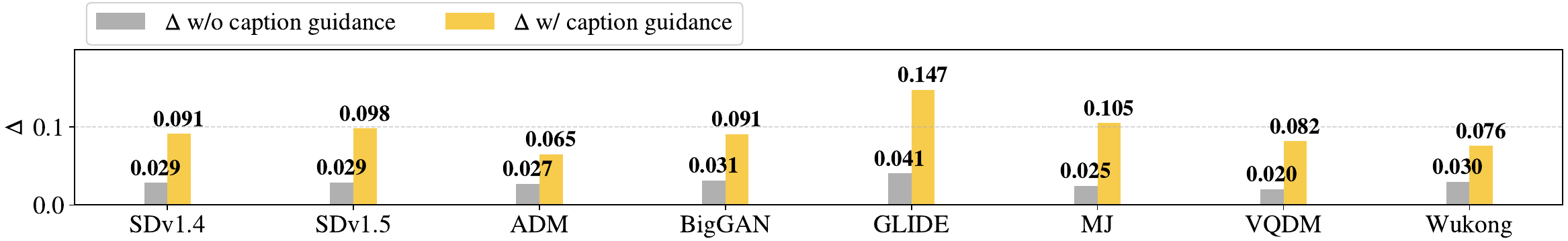}
        \caption{$\Delta$ values measuring the LPIPS score gap between real and fake images.}
        \label{delta_plot}
    \end{subfigure}
    \caption{Semantic shift analysis based on LPIPS scores~\cite{lpips}. Higher scores indicate lower similarity between the original and reconstructed images. Images are reconstructed under two conditions: with and without caption guidance.}
    \label{lpips}
\end{figure*}

\subsection{Cross-dataset Evaluation}
To further assess the generalization ability of the detection methods, we conducted a cross-dataset evaluation.
\cref{ood_acc} and \cref{ood_auc} report the accuracy and AUC score of each method on this test set.
Our method shows strong performance across diverse generative models, yielding an average accuracy of 71.47\% and an average AUC of 84.24\%, which are the highest among all detectors.
These results highlight the effectiveness of our method in OOD scenarios, demonstrating its robust generalization to fake images from unseen generative models.

\begin{figure*}[t]
  \centering
  \includegraphics[width=\linewidth]{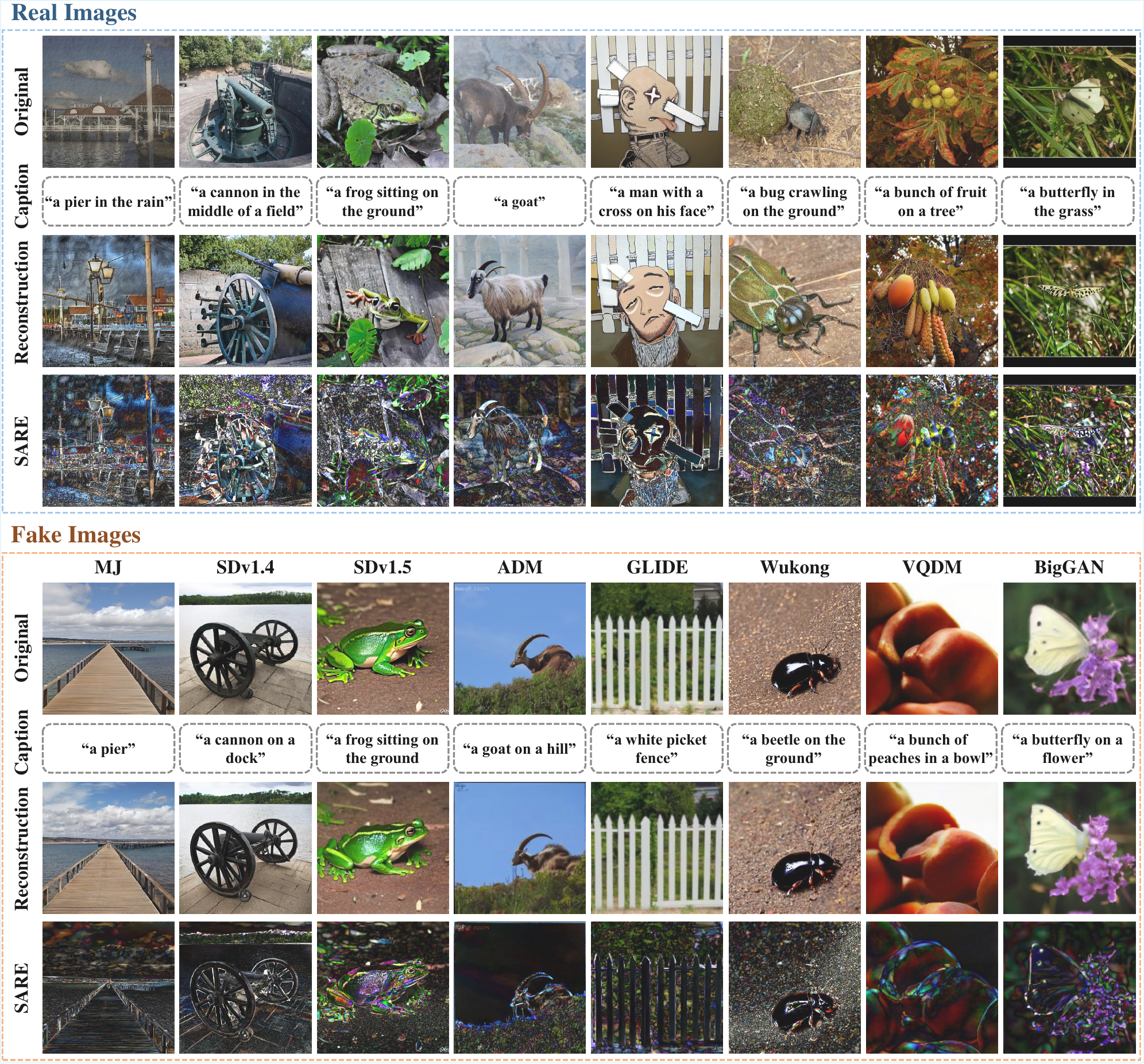}
\caption{Real and fake images from the GenImage dataset~\cite{genimage}, with captions generated by a pre-trained BLIP model~\cite{blip}, corresponding reconstructed images, and SAREs. The pixel values of the SARE are scaled by 2 for clearer visualization.}
  \label{fig:visualization}
\end{figure*}

\subsection{Semantic Shift Analysis}
\subsubsection{Quantitative Results.}
To validate the core assumption that real images undergo larger semantic shifts than fake images, we measured the perceptual distance between an image $x$ and its reconstruction $\hat{x}$ using the Learned Perceptual Image Patch Similarity (LPIPS)~\cite{lpips} metric.
\cref{lpips_plot} summarizes the average LPIPS scores for real and fake images in each subset of the GenImage dataset under two conditions: (1) reconstruction without caption guidance, and (2) reconstruction with caption guidance.
While real images consistently exhibit higher LPIPS scores than fake images in both settings, the gap between real and fake images is substantially larger when caption guidance is applied.
To quantify this gap, we define $\Delta$ as follows:
\begin{equation}
    \Delta = \mathbb{E}_{x \sim \mathcal{D}_{\text{real}}}[LPIPS(x, \hat{x})] 
           - \mathbb{E}_{x \sim \mathcal{D}_{\text{fake}}}[LPIPS(x, \hat{x})].
\label{delta}
\end{equation}
As shown in \cref{delta_plot}, $\Delta$ is relatively small without caption guidance, but increases significantly in all subsets when caption guidance is used.
These results suggest that the semantic difference between an image and its caption-guided reconstruction may serve as a more discriminative feature for detection, thereby leading to improved performance across diverse generative models.

\subsubsection{Qualitative Results and Visualizations.}

\cref{fig:visualization} presents qualitative examples of real and fake images from GenImage, along with their caption-guided reconstructions and the corresponding SAREs.
In GenImage, real images are sourced from ImageNet, while fake images are synthesized by generative models using ImageNet class labels as text prompts. 
For a fair comparison, we visualize real and fake images from the same ImageNet class label along with their reconstructions.
The results show that real images consistently exhibit larger semantic shifts than fake images.

\subsection{Ablation Study}
\subsubsection{Influence of Image Captioning Model.}
\label{ablation_caption_model}

\begin{table}[t]
\caption{Ablation study of different captioning models on the GenImage dataset~\cite{genimage}.}
\label{tab:ablation_caption}
\centering
\small
\setlength{\tabcolsep}{4pt}
\renewcommand{\arraystretch}{1.15}

\begin{tabularx}{\columnwidth}{
>{\centering\arraybackslash}m{1.38cm}
>{\centering\arraybackslash}m{2.85cm}
>{\centering\arraybackslash}m{0.9cm}
>{\centering\arraybackslash}m{1.15cm}
>{\centering\arraybackslash}m{1.15cm}
>{\centering\arraybackslash}m{0.9cm}
>{\centering\arraybackslash}m{0.9cm}
>{\centering\arraybackslash}m{0.9cm}
}
\toprule

\multirow{2}{*}[-0.6ex]{\shortstack{Captioning\\Model}} &
\multirow{2}{*}[-0.6ex]{Prompt} &
\multirow{2}{*}{\shortstack{Avg.\\Word\\Length}} &
\multirow{2}{*}[-0.6ex]{\shortstack{Avg.\\ACC(\%)}} &
\multirow{2}{*}[-0.6ex]{\shortstack{Avg.\\AUC(\%)}} &
\multicolumn{3}{c}{Avg. LPIPS Score} \\

\cmidrule(lr){6-8}

&
&
&
&
&
Real &
Fake &
$\Delta$ \\

\midrule

BLIP~\cite{blip}
& -
& 5.75
& 93.17
& 98.15
& 0.3437
& 0.2494
& 0.0943 \\

\midrule

\multirow[c]{3}{*}{\shortstack{LLaVA-\\NeXT~\cite{llavanext}}}
& Brief description within 50 words.
& 13.78
& 92.51
& 97.95
& 0.3407
& 0.2486
& 0.0921 \\

\cmidrule(lr){2-8}

& Detailed description within 80 words.
& 70.87
& 92.88
& 98.01
& 0.3406
& 0.2483
& 0.0923 \\

\bottomrule
\end{tabularx}

\end{table}

To evaluate the impact of different image captioning models on detection performance, we conducted an ablation study using captions generated by the pre-trained BLIP and LLaVA-NeXT-8B~\cite{llavanext} models on the GenImage dataset. 
We considered three captioning settings: (1) short captions generated by BLIP, (2) medium-length captions generated by LLaVA-NeXT using the prompt “Brief description within 50 words.”, and (3) long captions generated by LLaVA-NeXT using the prompt “Detailed description within 80 words.”
As shown in \cref{tab:ablation_caption}, BLIP produces concise captions with an average length of 5.75 words, whereas LLaVA-NeXT produces longer and more descriptive captions, averaging 13.78 and 70.87 words for the medium and long settings, respectively.
Despite these differences in caption length and descriptiveness, SARE achieves strong performance across all settings, indicating that its effectiveness does not depend on a specific captioning model.


We further measured the average LPIPS scores between the original and reconstructed images, as well as the gap between real and fake images defined in \cref{delta}. As presented in \cref{tab:ablation_caption}, real images consistently exhibit higher LPIPS scores than fake images across all settings. Although the absolute LPIPS values decrease for both real and fake images as captions become longer and more descriptive, the gap between them remains stable and is even larger in the long caption setting than in the medium length setting. These findings suggest that, even with more powerful captioning models, it remains challenging to fully capture the rich and fine-grained visual content of real images using natural language. As a result, caption-guided reconstruction introduces larger semantic changes for real images. In contrast, fake images tend to be more closely aligned with their captions and therefore undergo smaller semantic shifts. This observation supports our key hypothesis that the semantic shift captured by SARE reflects an intrinsic difference between real and fake images, regardless of caption length or descriptiveness. Qualitative examples of reconstructed images using BLIP and LLaVA-NeXT captions are provided in the supplementary material.

\begin{table*}[t]
\centering

\begin{minipage}[t]{0.48\textwidth}
\captionof{table}{Ablation study of different guidance scale values $w$ on the GenImage dataset~\cite{genimage}.}
\label{guidance}
\centering
\small
\begin{tabularx}{\linewidth}{
>{\centering\arraybackslash}m{1.8cm}|
>{\centering\arraybackslash}m{0.8cm}|
>{\centering\arraybackslash}m{1.4cm}
>{\centering\arraybackslash}m{1.4cm}
}
\toprule
Method & $w$
& \shortstack{Avg.\\ACC(\%)}
& \shortstack{Avg.\\AUC(\%)} \\
\midrule
DRCT & - & 88.81 & 95.66 \\
\midrule
\multirow{3}{*}{\shortstack{SARE\\(ours)}}
& 2.5  & 93.15 & 98.24 \\
& 7.5  & 93.17 & 98.15 \\
& 12.5 & 93.04 & 98.13 \\
\bottomrule
\end{tabularx}
\end{minipage}
\hfill
\begin{minipage}[t]{0.48\textwidth}
\captionof{table}{Ablation study of different fusion modules on the GenImage dataset~\cite{genimage}. CA denotes cross-attention.}
\label{fusion_module}
\centering
\small
\begin{tabularx}{\linewidth}{
>{\centering\arraybackslash}m{1.4cm}|
>{\centering\arraybackslash}m{1.45cm}|
>{\centering\arraybackslash}m{1.4cm}
>{\centering\arraybackslash}m{1.4cm}
}
\toprule
Method & \shortstack{Fusion\\Module}
& \shortstack{Avg.\\ACC(\%)}
& \shortstack{Avg.\\AUC(\%)} \\
\midrule
DRCT & - & 88.81 & 95.66 \\
\midrule
\multirow{3}{*}{\shortstack{SARE\\(ours)}}
& Concat & 91.81 & 97.53 \\
& FiLM~\cite{film} & 89.77 & 98.12 \\
& CA & 93.17 & 98.15 \\
\bottomrule
\end{tabularx}
\end{minipage}

\end{table*}

\subsubsection{Influence of Guidance Scale.}
We investigated the impact of the guidance scale $w$ on detection performance within the caption-guided reconstruction framework.
\cref{guidance} reports accuracy and AUC on GenImage for different guidance scale values.
The results show that incorporating SARE consistently improves the performance over the baseline across all settings.
The best accuracy is achieved at $w=7.5$ (93.17\%), whereas the highest AUC is observed at $w=2.5$ (98.24\%).

\subsubsection{Influence of Fusion Module.}
We conducted an ablation study to examine the influence of different fusion designs on integrating SARE into the backbone detector.
For Concat, image and semantic features are concatenated along the channel dimension.
For FiLM~\cite{film}, semantic features generate scaling and shifting parameters to modulate image features via feature-wise affine transformation.
As reported in \cref{fusion_module}, all fusion strategies consistently outperform the baseline, confirming the effectiveness of the semantic information provided by SARE.
Cross-attention achieves the best performance, indicating that adaptively attending to semantic representations enables effective integration of SARE into the detection framework.

\begin{table}[t]

\caption{Accuracy (ACC, \%) comparisons of different detectors in robustness evaluation on the GenImage dataset~\cite{genimage}. QF denotes JPEG quality factor.}
\label{robustness_acc}

\centering
\small
\setlength{\tabcolsep}{4pt}
\resizebox{\textwidth}{!}{
\begin{tabular}{lccccc}
\toprule
Method & JPEG (QF=90) & JPEG (QF=80) & JPEG (QF=70) & Scale (0.75) & Scale (1.25) \\
\midrule
GramNet~\cite{gramnet} & 71.22 & 71.02 & 71.22 & 69.89 & 66.31 \\
Conv-B~\cite{convb}       & 71.91 & 71.57 & 71.42 & 75.30 & 75.28 \\
UnivFD~\cite{univfd}        & 73.24 & 70.42 & 69.75 & 75.96 & 73.99 \\
DIRE~\cite{dire}      & 52.79 & 50.65 & 50.34 & 51.94 & 53.77 \\
DE-FAKE~\cite{defake}         & 71.00 & 70.88 & 70.44 & 72.33 & 71.88 \\
NPR~\cite{npr}       & 71.10 & 70.71 & 70.07 & 72.22 & 71.54 \\
AIDE~\cite{aide}    & 57.22 & 58.50 & 60.45 & 83.13 & 83.79 \\
\cmidrule(lr){1-6}
DRCT~\cite{drct}        & 80.97 & 78.06 & 76.18 & 79.51 & 74.75 \\
\rowcolor{gray!15}
SARE (ours) & 85.64 & 82.72 & 79.14 & 87.60 & 82.74 \\
\bottomrule
\end{tabular}
}

\end{table}

\begin{table}[t]

\caption{AUC (\%) comparisons of different detectors in robustness evaluation on the GenImage dataset~\cite{genimage}. QF denotes JPEG quality factor.}
\label{robustness_auc}

\centering
\small
\setlength{\tabcolsep}{4pt}
\resizebox{\textwidth}{!}{
\begin{tabular}{lccccc}
\toprule
Method & JPEG (QF=90) & JPEG (QF=80) & JPEG (QF=70) & Scale (0.75) & Scale (1.25) \\
\midrule
GramNet~\cite{gramnet} & 80.12 & 81.71 & 81.15 & 69.47 & 69.32 \\
Conv-B~\cite{convb}       & 90.53 & 90.17 & 90.43 & 94.15 & 96.90 \\
UnivFD~\cite{univfd}        & 85.31 & 82.53 & 81.08 & 85.48 & 82.05 \\
DIRE~\cite{dire}      & 70.83 & 63.44 & 59.23 & 78.92 & 80.61 \\
DE-FAKE~\cite{defake}         & 79.89 & 79.02 & 77.96 & 81.09 & 78.10 \\
NPR~\cite{npr}       & 76.87 & 76.50 & 76.57 & 91.20 & 92.32 \\
AIDE~\cite{aide}    & 75.23 & 78.12 & 80.66 & 96.28 & 96.00 \\
\cmidrule(lr){1-6}
DRCT~\cite{drct}        & 88.89 & 86.18 & 84.05 & 89.29 & 87.93 \\
\rowcolor{gray!15}
SARE (ours) & 94.33 & 93.10 & 89.68 & 94.97 & 92.77 \\
\bottomrule
\end{tabular}
}

\end{table}

\subsection{Robustness to Perturbations}
We evaluated the robustness of the detectors against two types of post-processing perturbations: JPEG compression (quality factors of 90, 80, and 70) and image resizing (scales of 0.75 and 1.25).
\cref{robustness_acc} and \cref{robustness_auc} report the average accuracies and AUC scores of different detection methods on the GenImage dataset under these settings.
Except for NPR and AIDE, all models were trained using the same data augmentations. 
Our method consistently outperforms DRCT and maintains superior performance across all perturbation conditions, suggesting that incorporating SARE improves robustness to post-processing distortions.
\section{Conclusion}

We introduced a novel representation for AI-generated image detection, termed Semantic-Aware Reconstruction Error (SARE), which quantifies the semantic difference between an image and its caption-guided reconstruction.
By leveraging the relationship between an image and its caption, SARE provided a discriminative and generalizable feature for detecting fake images across diverse generative models.
Experimental results showed that SARE significantly improved detection performance in both ID and OOD settings, surpassing existing baselines.

\bibliographystyle{splncs04}
\bibliography{main}

\appendix
\section*{Appendix}
\addcontentsline{toc}{section}{Appendix}

\section{Additional Ablation Studies}
\subsection{Influence of \textit{Strength} Parameter}
To evaluate the influence of the \textit{strength} parameter on detection performance, we conducted an ablation study on the GenImage~\cite{genimage} dataset by varying the \textit{strength} value from 0.4 to 0.8.
\cref{strength} shows the accuracy and AUC performance for each \textit{strength} value.
The results demonstrate that the model maintains stable performance across all values.
The highest accuracy is obtained at \textit{strength} = 0.5, while the best AUC is achieved at \textit{strength} = 0.7.

\begin{table}[t]

\caption{Ablation study of the \textit{strength} parameter on the GenImage dataset~\cite{genimage}.}
\label{strength}

\centering
\small
\begin{tabular}{>{\centering\arraybackslash}p{1.5cm}|cc}
\toprule
$strength$ & Avg. ACC(\%) & Avg. AUC(\%) \\
\midrule
0.4  & 92.61 & 97.97 \\
0.5  & 93.17 & 98.15 \\
0.6 & 92.74 & 98.06 \\
0.7 & 93.16 & 98.19 \\
0.8 & 92.98 & 98.18 \\
\bottomrule
\end{tabular}

\end{table}

\begin{table}[t]

\caption{Ablation study of semantic encoders on the GenImage dataset~\cite{genimage}.}
\label{semantic_encoder}

\centering
\small
\begin{tabular}{p{2.9cm}|cc}
\toprule
Semantic Encoder & Avg. ACC(\%) & Avg. AUC(\%) \\
\midrule
EfficientNet-B3~\cite{effinet} & 93.40 & 98.35 \\
ConvNeXt-Base~\cite{convb}   & 92.72 & 98.17 \\
ResNet50~\cite{resnet}        & 93.17 & 98.15 \\
\bottomrule
\end{tabular}

\end{table}

\begin{table}[t]

\caption{Ablation study of reconstruction models on the GenImage dataset~\cite{genimage}.}
\label{recon_model}

\centering
\small
\begin{tabular}{p{3.2cm}|cc}
\toprule
Reconstruction Model & Avg. ACC(\%) & Avg. AUC(\%) \\
\midrule
SDv1~\cite{sd}      & 93.17 & 98.15 \\
SDXL~\cite{sdxl}      & 92.22 & 97.77 \\
SwiftEdit~\cite{swiftedit} & 91.50 & 97.65 \\
\bottomrule
\end{tabular}

\end{table}

\begin{table}[t]

\caption{Ablation study of backbone detectors on the GenImage dataset~\cite{genimage}.}
\label{sare_backbone}

\centering
\small
\begin{tabular}{p{2.4cm}|cc}
\toprule
Method & Avg. ACC(\%) & Avg. AUC(\%) \\
\midrule
ResNet50~\cite{resnet}          & 73.52 & 91.87 \\
\textbf{+} SARE (ours) & 75.37 & 95.29 \\
\midrule
UnivFD~\cite{univfd}        & 79.11 & 90.61 \\
\textbf{+} SARE (ours) & 81.44 & 95.20 \\
\midrule
DRCT~\cite{drct}          & 88.81 & 95.66 \\
\textbf{+} SARE (ours) & 93.17 & 98.15 \\
\bottomrule
\end{tabular}

\end{table}

\begin{table}[t]
\caption{Comparison of different detectors on the GenImage dataset~\cite{genimage}. We analyze the use of captioning and reconstruction models, inference time (seconds per image), and detection performance. In SARE, SDv1~\cite{sd} or SwiftEdit~\cite{swiftedit} is used as the reconstruction models.}
\label{tab:inference}
\centering
\small
\begin{tabular}{c c c c c c}
\toprule
Method 
& \makecell{Caption. \\ Model} 
& \makecell{Recon. \\ Model} 
& \makecell{Avg. Inference \\ Time(s/img)} 
& \makecell{Avg. \\ ACC(\%)} 
& \makecell{Avg. \\ AUC(\%)} \\
\midrule
DE-FAKE~\cite{defake} 
& \cmark & \xmark 
& 0.17 & 74.60 & 86.44 \\
DIRE~\cite{dire} 
& \xmark & \cmark 
& 2.68 & 70.35 & 83.01 \\
AEROBLADE~\cite{aeroblade} 
& \xmark & \cmark 
& 0.90 & 80.17 & 86.15 \\
SARE\textbf{/}SDv1 (ours)
& \cmark & \cmark 
& 1.60 & 93.17 & 98.15 \\
SARE\textbf{/}SwiftEdit (ours)
& \cmark & \cmark 
& 0.80 & 91.50 & 97.65 \\
\bottomrule
\end{tabular}
\end{table}

\subsection{Influence of Semantic Encoder}
We evaluated three different architectures for the semantic encoder. As presented in \cref{semantic_encoder}, SARE consistently achieves strong performance across all settings on the GenImage dataset, indicating that the effectiveness of SARE does not depend on a specific architecture for semantic feature extraction.

\subsection{Influence of Reconstruction Model}
To evaluate the influence of the reconstruction model on detection performance, we conducted an ablation study using three different reconstruction models. While SDv1~\cite{sd} and SDXL~\cite{sdxl} perform caption-guided reconstruction through iterative diffusion steps, SwiftEdit~\cite{swiftedit} requires only a single denoising step, substantially reducing the computational cost of the reconstruction process. \cref{recon_model} shows that SARE maintains stable performance across all reconstruction models.

\subsection{Influence of Backbone Detector}
We evaluated SARE with three backbone detectors, and the results on the GenImage dataset are summarized in \cref{sare_backbone}. In all cases, incorporating SARE consistently improves both accuracy and AUC compared to the backbone models.
These results suggest that SARE provides robust and generalizable semantic cues that enhance detection performance regardless of the backbone architecture.

\section{Additional Visualizations}
\label{app_visual}
\begin{figure}[p]
    \centering
    \includegraphics[width=0.85\linewidth]{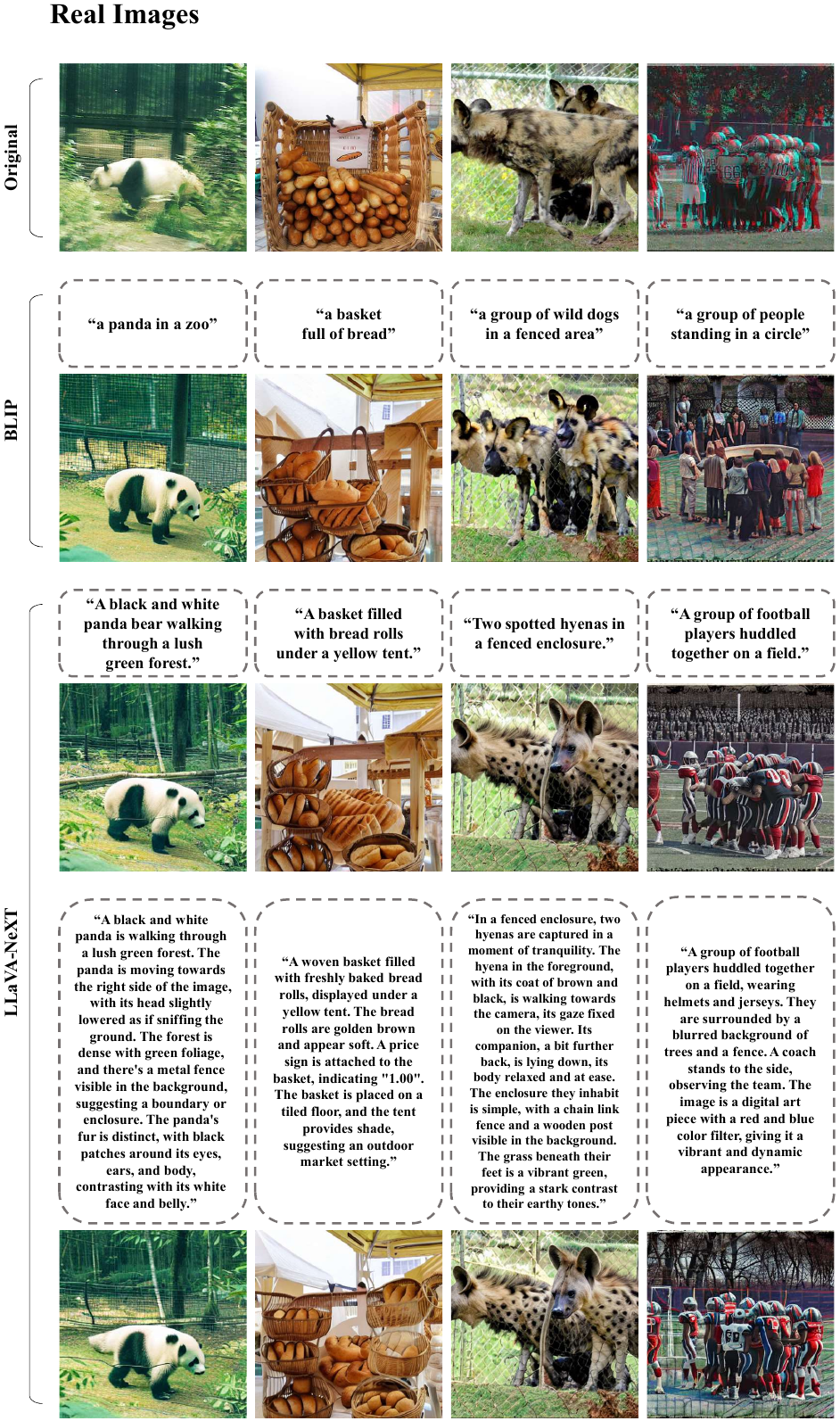}
    \caption{Real images from the GenImage dataset~\cite{genimage} with captions from BLIP~\cite{blip} and LLaVA-NeXT~\cite{llavanext}, and their corresponding reconstructions. For LLaVA-NeXT, we used the prompts “Brief description within 50 words.” and “Detailed description within 80 words.”}
    \label{fig:llava_visual_1}
\end{figure}

\begin{figure}[p]
    \centering
    \includegraphics[width=0.85\linewidth]{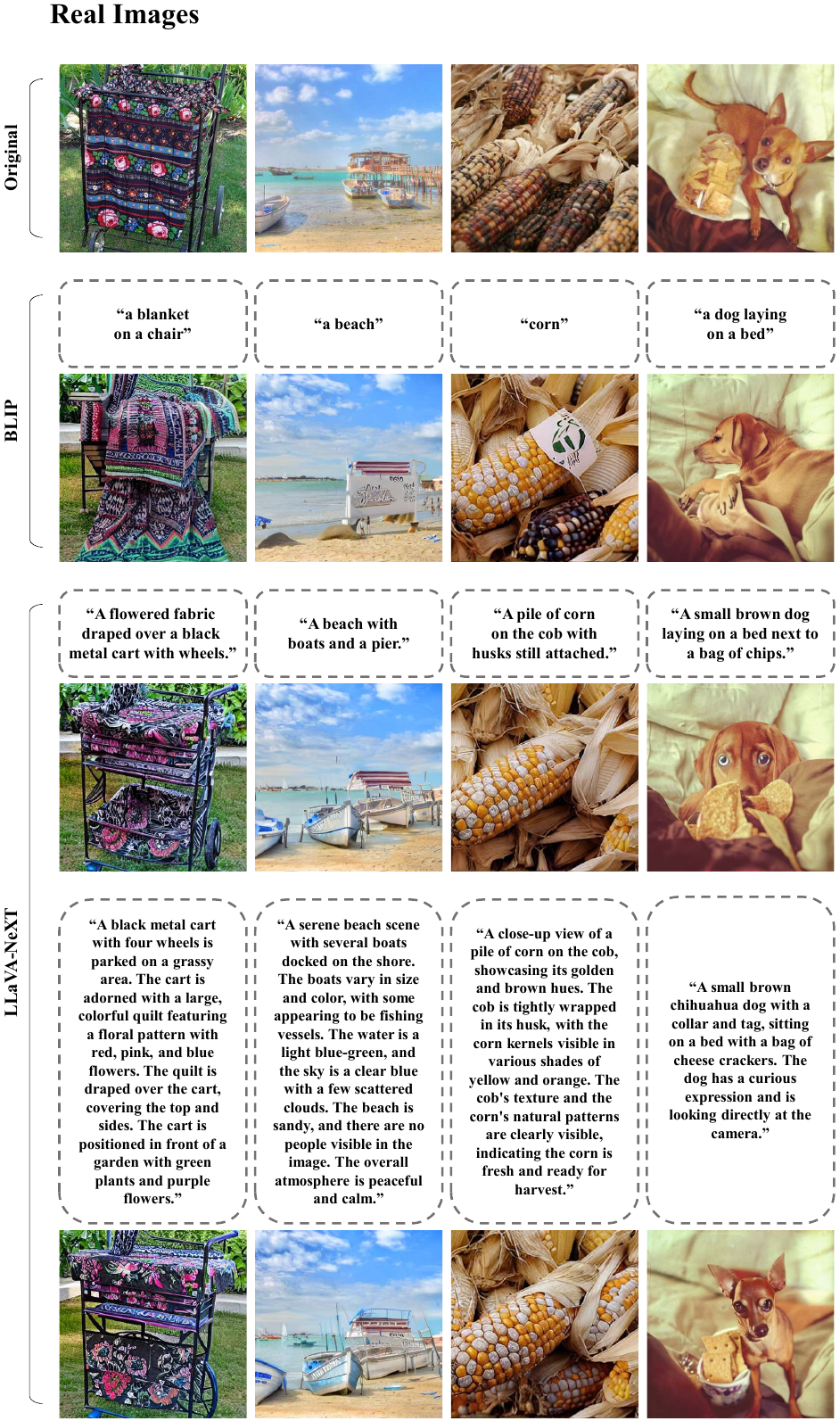}
    \caption{Real images from the GenImage dataset~\cite{genimage} with captions from BLIP~\cite{blip} and LLaVA-NeXT~\cite{llavanext}, and their corresponding reconstructions. For LLaVA-NeXT, we used the prompts “Brief description within 50 words.” and “Detailed description within 80 words.”}
    \label{fig:llava_visual_2}
\end{figure}

\begin{figure}[p]
    \centering
    \includegraphics[width=0.85\linewidth]{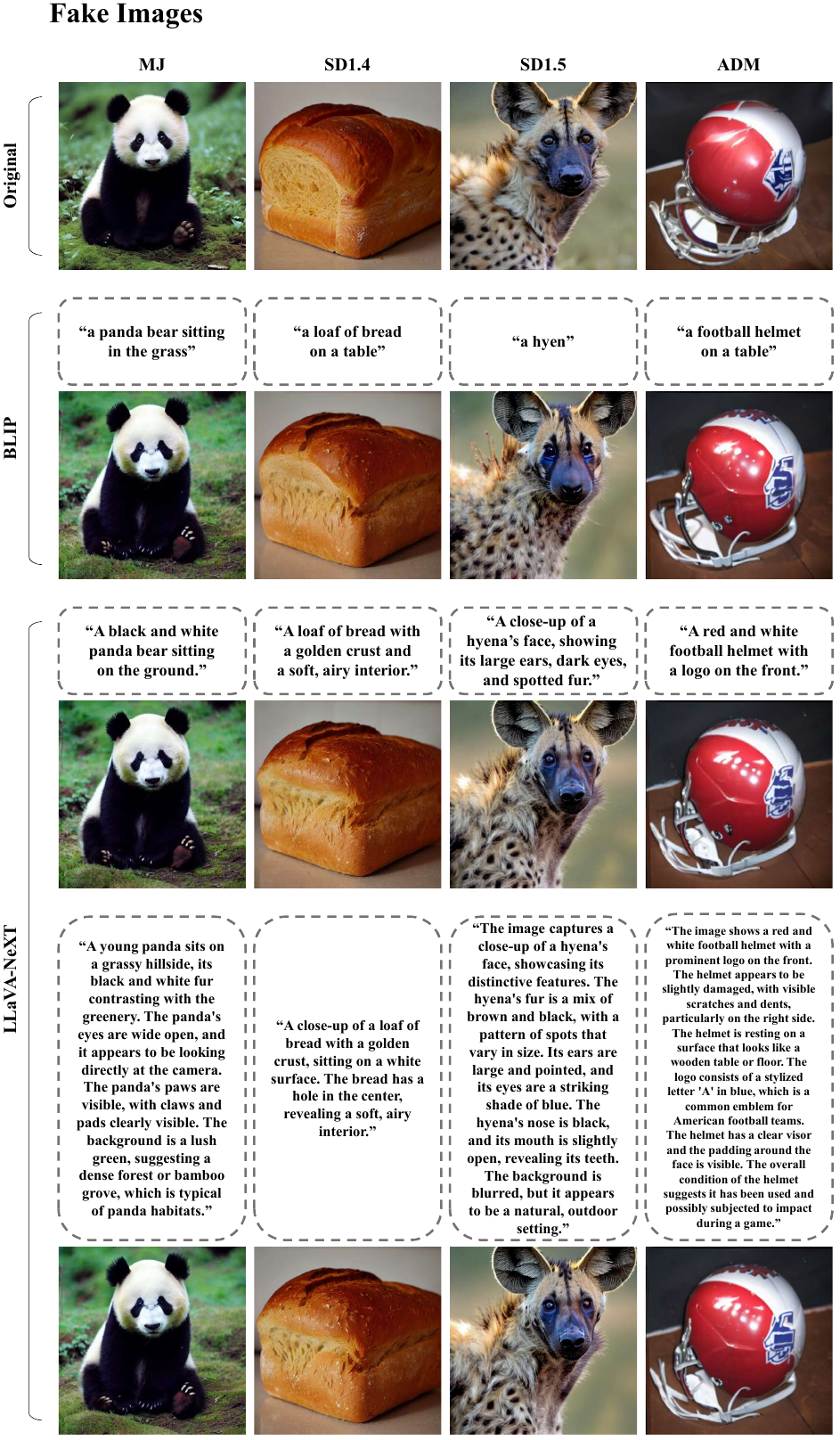}
    \caption{Fake images from the GenImage dataset~\cite{genimage} with captions from BLIP~\cite{blip} and LLaVA-NeXT~\cite{llavanext}, and their corresponding reconstructions. For LLaVA-NeXT, we used the prompts “Brief description within 50 words.” and “Detailed description within 80 words.”}
    \label{fig:llava_visual_3}
\end{figure}

\begin{figure}[p]
    \centering
    \includegraphics[width=0.85\linewidth]{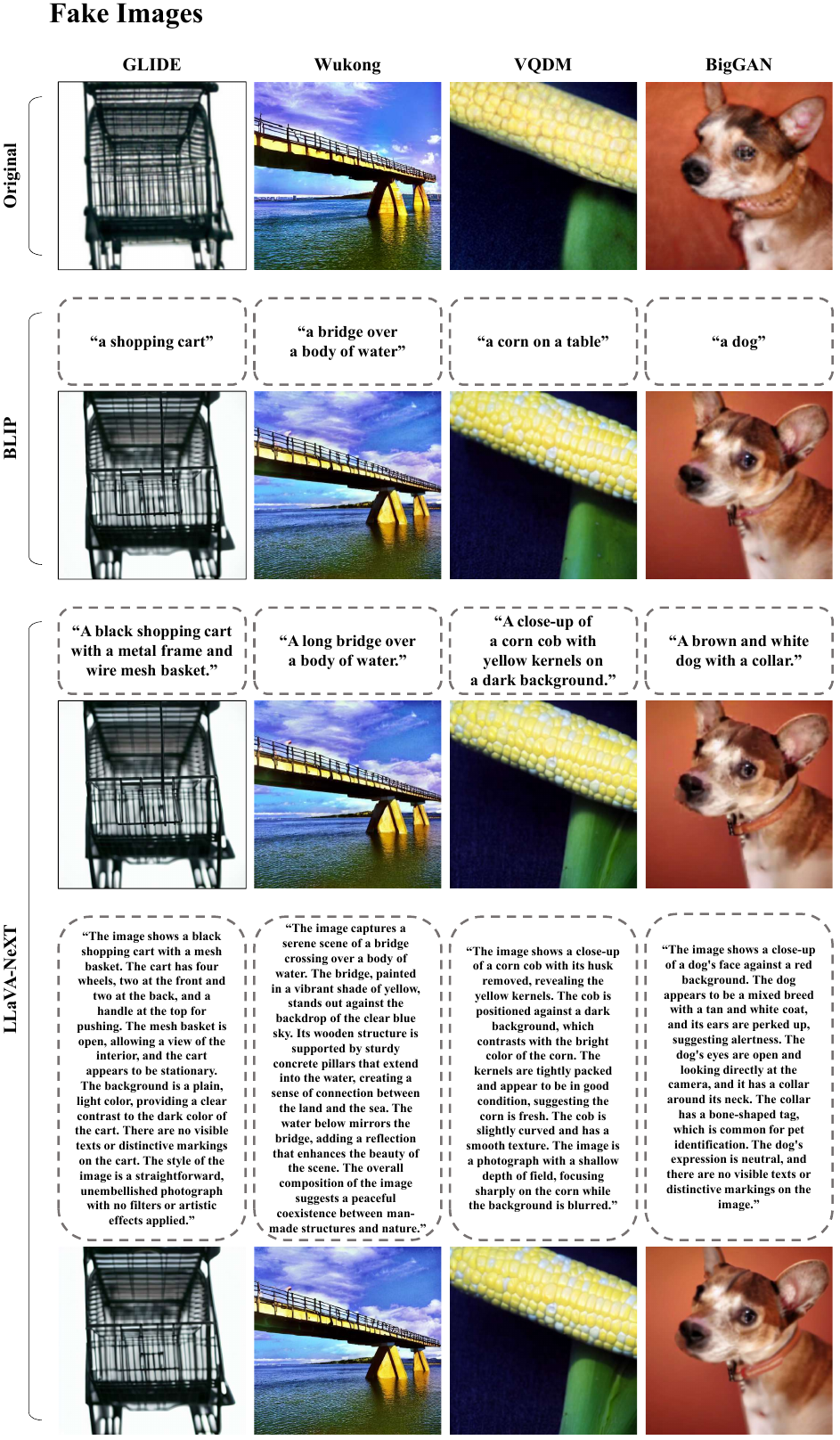}
    \caption{Fake images from the GenImage dataset~\cite{genimage} with captions from BLIP~\cite{blip} and LLaVA-NeXT~\cite{llavanext}, and their corresponding reconstructions. For LLaVA-NeXT, we used the prompts “Brief description within 50 words.” and “Detailed description within 80 words.”}
    \label{fig:llava_visual_4}
\end{figure}

\cref{fig:llava_visual_1,fig:llava_visual_2,fig:llava_visual_3,fig:llava_visual_4} present real and fake images from the GenImage dataset and their reconstructions guided by captions from BLIP~\cite{blip} and LLaVA-NeXT~\cite{llavanext}. To ensure a fair comparison, real and fake images are selected from the same ImageNet~\cite{imagenet} class label. The visualizations show that real images typically undergo larger semantic shifts than fake images during caption-guided reconstruction with both captioning models.

\section{Limitation}

Although SARE achieves strong performance, its caption-guided reconstruction framework introduces additional computational overhead due to the image captioning and diffusion-based reconstruction processes. To analyze this overhead, we compare inference time (seconds per image) and detection performance across caption-based and reconstruction-based detectors in \cref{tab:inference}.
Inference times were measured on a single NVIDIA GeForce RTX 3090 Ti.
As shown in \cref{tab:inference}, the caption-based method DE-FAKE~\cite{defake} exhibits substantially lower inference time than reconstruction-based approaches, indicating that the computational burden of caption generation is relatively minor compared to that of image reconstruction.
Reconstruction-based detectors typically require high computational cost, primarily due to the iterative denoising process.
The longer inference times of DIRE~\cite{dire} and SARE/SDv1 arise from multi-step denoising, whereas AEROBLADE~\cite{aeroblade} reduces computational cost by leveraging an autoencoder-based reconstruction.
Accordingly, the computational cost of SARE can be effectively mitigated by employing SwiftEdit~\cite{swiftedit}, which requires only a single denoising step.
SARE/SwiftEdit achieves the lowest runtime among reconstruction-based detectors, despite incorporating the additional captioning module.
Given the substantial improvements in detection performance, we believe that the additional overhead introduced by SARE remains practically acceptable.
The results demonstrate that SARE achieves a balanced trade-off between computational efficiency and detection performance.


\clearpage

\end{document}